\documentclass[letterpaper, 10 pt, conference]{ieeeconf}

\IEEEoverridecommandlockouts                              

\usepackage{amsmath,amssymb,amsfonts}
\usepackage{algorithm}
\usepackage{algpseudocode}
\usepackage{caption}
\usepackage{subcaption}
\usepackage{ulem}
\usepackage{hyperref}
\usepackage{tabularx}
\usepackage{booktabs}
\usepackage{bm}
\usepackage{graphicx}
\graphicspath{{images/}}

\makeatletter
\newcommand\fs@betterruled{%
  \def\@fs@cfont{\bfseries}\let\@fs@capt\floatc@ruled
  \def\@fs@pre{\vspace*{5pt}\hrule height.8pt depth0pt \kern2pt}%
  \def\@fs@post{\kern2pt\hrule\relax}%
  \def\@fs@mid{\kern2pt\hrule\kern2pt}%
  \let\@fs@iftopcapt\iftrue}
\floatstyle{betterruled}
\restylefloat{algorithm}
\makeatother

\usepackage[dvipsnames]{xcolor}

\begin{document}


\title{\LARGE \bf A Human-Centered Dynamic Scheduling Architecture for Collaborative Application$^*$}


%
%

\author{Andrea Pupa$^1$, Wietse Van Dijk$^2$ and Cristian Secchi$^1$
    \thanks{$^*$ This project has received funding from the European Union’s Horizon 2020 research and innovation programme under grant agreement No. 818087 (ROSSINI).}
	\thanks{$^{1}$ Andrea Pupa and Cristian Secchi are with the Department of Science and Method of Engineering, University of Modena and Reggio Emilia, Italy. E-mail:
		{\tt\small \{\href{mailto:andrea.pupa@unimore.it}{andrea.pupa}, \href{mailto:cristian.secchi@unimore.it}{cristian.secchi}\}@unimore.it}}%
	\thanks{$^{2}$ Wietse Van Dijk is with the Netherlands Organisation for applied scientific research - TNO, Holland. E-mail:
		{\tt\small \href{mailto:wietse.vandijk@tno.nl}{wietse.vandijk@tno.nl}}}%
}


\maketitle

\begin{abstract}
In collaborative robotic applications, human and robot have to work
together during a whole shift for executing a sequence of jobs. The performance of the human robot team can be enhanced by scheduling the right tasks to the human and the robot. The scheduling should consider the task execution constraints, the variability in the task execution by the human, and the job quality of the human. Therefore, it is necessary to dynamically schedule the assigned tasks. In
this paper, we propose a two-layered architecture for task allocation
and scheduling in a collaborative cell. Job quality is
explicitly considered during the allocation of the tasks and over a
sequence of jobs. The tasks are dynamically scheduled based on the
real time monitoring of the human's activities. The effectiveness of
the proposed architecture is  experimentally validated.

\end{abstract}

\section{Introduction} 

In recent years, industrial setting has been supported by a constant increase in the use of collaborative robotics (see e.g. \cite{villani2018, bauer2008}).

The shift towards collaborative robotics can significantly change the quality of the job for the human. In fact, collaborative robots can take over dull, heavy or dangerous tasks making the life of the human easier. To ensure that the re-distribution of tasks is favorable to the human, the job quality aspect has to be taken into account when dividing tasks between human and robot \cite{pham2018impact}. This process can be guided by using job-quality metrics. The job quality framework, as used by the OECD \cite{oecd2016}, is a multidimensional concept that covers various topics. The part of job quality that concerns the quality of the working environment (e.g. time pressure, physical risks and work autonomy) is of specific interest for human-robot collaboration. At an even lower level, aspects of task load are governed by guidelines and regulations, for example for lifting loads \cite{health1981} and noise exposure \cite{eudirective}.

In the industrial scenarios, collaborative cells are built in order to enable the human and the robot to work together on various jobs,  each  of  which  composed  by  a  set  of  tasks. The distribution of task determines the load to which the human is subjected (\cite{pham2018impact,wixted2014effect}). Additionally, the distribution of tasks determines influences the fluency, which in turn is strongly related to job quality aspects \cite{hoffman2019evaluating}.

A lot of research has been done in the multi-agent task allocation problem in the industrial cases (see e.g. \cite{blum2011,xu2018,sabattini2015}). In general,  these solutions cannot be directly applied in a human-robot collaboration (HRC) application, as they consider the presence of homogeneous agents.

Task allocation for collaborative cells has been modeled as a nonlinear optimization problem (see e.g. \cite{George2018,Kai2019,Ashkan2020, bogner2018}) but the  computational complexity of the problem is often high and it does not explicitly allow to take into account variable job-quality parameters. In \cite{rahman2018} a two-level feedforward optimization strategy for offline subtask allocation between human and robot is presented. This strategy is integrated with a feedback procedure based on mutual trust to re-allocate the subtasks online. In \cite{nikolakis2018} the authors propose a multi-criteria decision-making framework for task allocation which generates a solution that best matches the criteria you want to optimize. Moreover, in case of unexpected events, the algorithm can be exploited for re-scheduling the remaining tasks. In \cite{johannsmeier2016} a two-level framework for task assignment is presented that dynamically handles task failures in a HRC scenario. Nevertheless, job quality over several jobs is not considered. Task assignment strategies for HRC assume that both the human and the robot require a constant amount of time to perform a task. This assumption may lead to inefficiencies. In reality, the human does not always take the same amount of time to accomplish the same task. Several works that consider task rescheduling with a variable human execution time are available in the literature (see e.g. \cite{lou2012,casalino2019}) but the operator and the robot are treated as two separate entities and human-robot interaction and communication is not considered. 

In order to make the human-robot collaboration as natural as possible, it is necessary to improve mutual awareness and communication both at the task execution level and at the task planning and scheduling level. Thus, it is important to have a monitoring strategy that makes the scheduler aware of the real duration of the tasks executed by the human and about the job quality of the human. This strategy allows the scheduler to adapt the assigned tasks improving the job quality and the efficiency of the human-robot collaboration. Moreover, both the human and the robot should be able to communicate through the scheduler in order to improve the collaboration and to make it as natural as possible. The human, due to its expertise and experience, should be able to decide to execute a task that was previously assigned to the robot. While the robot should be able to assign a specific task to the human, if some failure occurs (e.g. a robot tool broke). All these decisions should be handled by the scheduler which re-assigns the tasks accordingly.

In this paper we propose novel framework for task assignment and scheduling for collaborative cells that is aware about the activity of the human and that allows the human and the robot to take decisions about the tasks they need to execute. This framework intrinsically considers job quality in the scheduling algorithm, in order to improve job quality of human workers.  The proposed framework is composed by two layers.
The first layer assigns, off-line, the tasks within a job to either the human or the robot by providing a nominal schedule. It considers the actual job quality indexes and the dependencies between the tasks. The second layer, the scheduler layer, reschedules the tasks considering the real execution time of the human operator, if needed, and the decisions taken online both by the human and the robot.

The main contributions of this paper are:
\begin{itemize}
	\item A novel adaptive framework for task assignment and scheduling that considers into account real execution time, the job quality of the human, and the communication with human and robot for dynamic rescheduling 
	\item A strategy for dynamic rescheduling that is effective and computationally cheap, i.e. suitable for industrial applications, and that allows human and robot to communicate their needs to the scheduler. 
\end{itemize}

 The paper is organized as follows: Sec.~\ref{sec:problem} presents the task assignment and dynamic scheduling problem for a collaborative cell. Sec.~\ref{sec:architecture} presents the overall proposed architecture. Sec.~\ref{sec:task_planning} an optimization problem for solving the task assignment and scheduling considering job quality is proposed. Sec.~\ref{sec:job_quality} defines the quantitative job quality metrics and attractiveness factor for human-robot collaboration. Sec.~\ref{sec:dynamic_scheduling} presents an algorithm that dynamically schedules the tasks for the human and the robot. Sec.~\ref{sec:experiments} summarizes the experimental validation of the proposed architecture. Sec.~\ref{sec:conclusions} sums the conclusions and proposes future work.

\section{Problem Statement} 
\label{sec:problem}
A collaborative industrial workspace is characterized by the presence of two different agents, a human operator $H$ and a robot $R$, that must cooperate during a work shift in order to perform $S$ jobs $(J_1,\dots ,J_S)$. Each job consists out of one or more tasks\footnote{The choice of the specific technique for splitting a job into several tasks is out of the scope of this paper. Several strategies are available in the literature (see, e.g., \cite{johannsmeier2016} for assembly tasks.)} $(T_{1_j},\dots ,T_{N_j})$, each of which is characterized by an intrinsic cost $w_{Ai}$ and by a \emph{nominal} execution time $t_{ai}$, where $a \in A = \{H,R\}$ represents the agent that executes the task $i$. In the rest of the paper the double index is removed for ease of notation and we refer to the set of tasks $(T_{1_j},\dots ,T_{N_j})$ as $(T_{1},\dots ,T_{N})$. 
The real task execution time can differ from the nominal execution time, due to uncertainties in the human behavior. Therefore, the workspace is equipped with a monitoring unit that, for each task $T_i$ assigned to the human, estimates online the real execution time. To achieve this, many solutions can be found in literature, e.g. sequential interval networks \cite{vo2014}, interaction  probabilistic movement primitives \cite{maeda2017}, Open-Ended Dynamic Time Warping (OE-DTW) \cite{maderna2019}.
The output from the monitoring unit is used to update all relevant parameters over the entire work-shift.

The tasks composing a job may depend on each other, i.e. there could be precedence constraints, and the execution order must be considered when assigning tasks.

In this work, we aim at designing a task assignment and dynamic scheduling architecture that:
\begin{itemize}
	\item Builds optimal nominal task schedules for the human and the robot, i.e. two task schedules such that, considering the nominal execution times, the precedence constraints, and job quality metrics, minimizes the job makespan, maximizing the parallelism between human and robot (i.e. minimizes waiting time), and optimizing the job quality for the human operator over the entire work shift.
	\item Starting from the nominal task schedules, reschedules both the human and the robot tasks according to the real execution time detected by the monitoring unit and the decisions taken by the human and the robot for task swapping. The rescheduling aims at minimizing the job makespan and improving the collaboration between the two agents. 
\end{itemize}

\section{Architecture} 
\label{sec:architecture}
The proposed task assignment and dynamic scheduling strategy is shown in Fig.~\ref{fig:framework}, where two main layers can be distinguished:
\begin{figure}[t]
	\centerline{\includegraphics[width=\columnwidth]{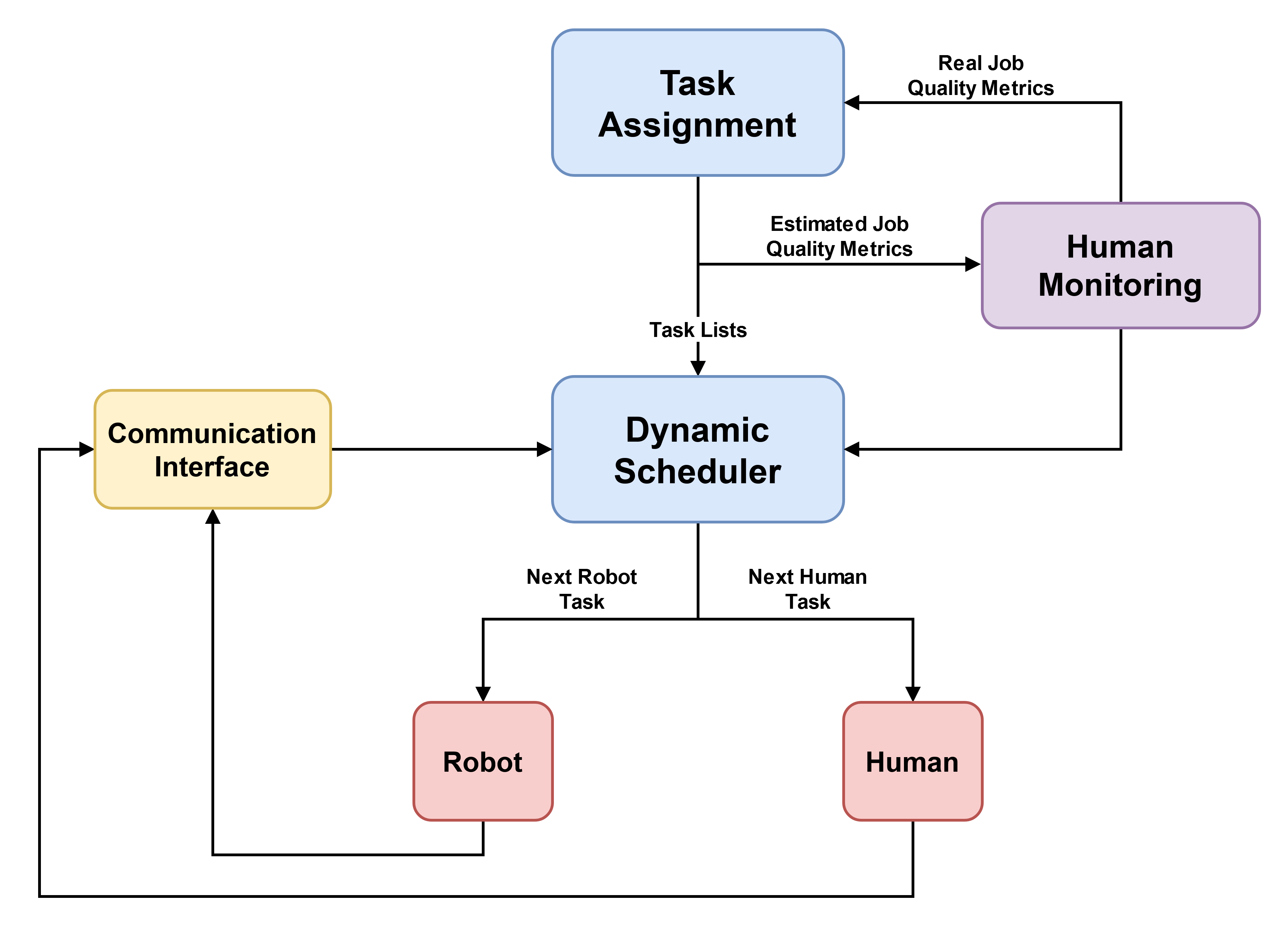}}
	\caption{The overall architecture. The blue blocks represent the two layers. The yellow and the purple blocks symbolize the strategies implemented to provide richer information to the architecture. The red blocks represent the two agents.} 
	\label{fig:framework}
\end{figure}
\begin{enumerate}
	\item \textbf{The Task Assignment Layer} is responsible of generating initial nominal schedules for the robot and the human, based on the maximum parallelism criterion, taking into account precedence constraints and job quality metrics over the entire work shift. 
	\item \textbf{The Dynamic Scheduler Layer} is responsible of scheduling the tasks, considering the real execution time and the requests coming from the human and from the robot.
\end{enumerate}

Once the Task Assignment layer computes the initial nominal schedules, an estimation of the human job quality parameters is performed. This estimation is based on the nominal execution time $t_{Hi}$.

The resulting estimate represents the expected job quality parameters if the behavior of the human was exactly as the nominal one. To accommodate for (expected) deviations from the nominal schedule, the estimated job quality parameters are given as input to the \textbf{Human Monitoring} block, which is responsible for supporting and improving the scheduling procedure. 

The Human Monitoring block aims to track the real execution time of the human operator during the execution of assigned tasks. The information about the real human behavior is then leveraged to update the estimated parameters. Subsequently, the real parameters that come out of the Human Monitoring block are used as input for the Task Assignment layer when calculating the nominal schedules of the new job.
This feedback procedure allows to keep track of the evolution of job quality throughout the entire work shift, adapting each schedule accordingly.

The real execution time of the human is also exploited by the Dynamic Scheduler, which aims to reschedule, in real-time, the nominal tasks schedules. Frequently changing the order of the tasks assigned to the human can lead to confusion and poor efficiency of the human \cite{Sandra2002}. Thus, we have chosen to focus the rescheduling strategy primarily robot tasks. The list of tasks assigned to the human changes only when required by the \textbf{Communication Interface} block, namely when the robot cannot execute a task and a failure occurs, and when the human decides to perform a task instead of the robot. In all these cases, the changes in the human schedule are necessary and minimal.
\section{Task Assignment} 
\label{sec:task_planning}
\begin{figure}[t]
  \centering
  \subfloat[]{\includegraphics[height=0.2\textheight]{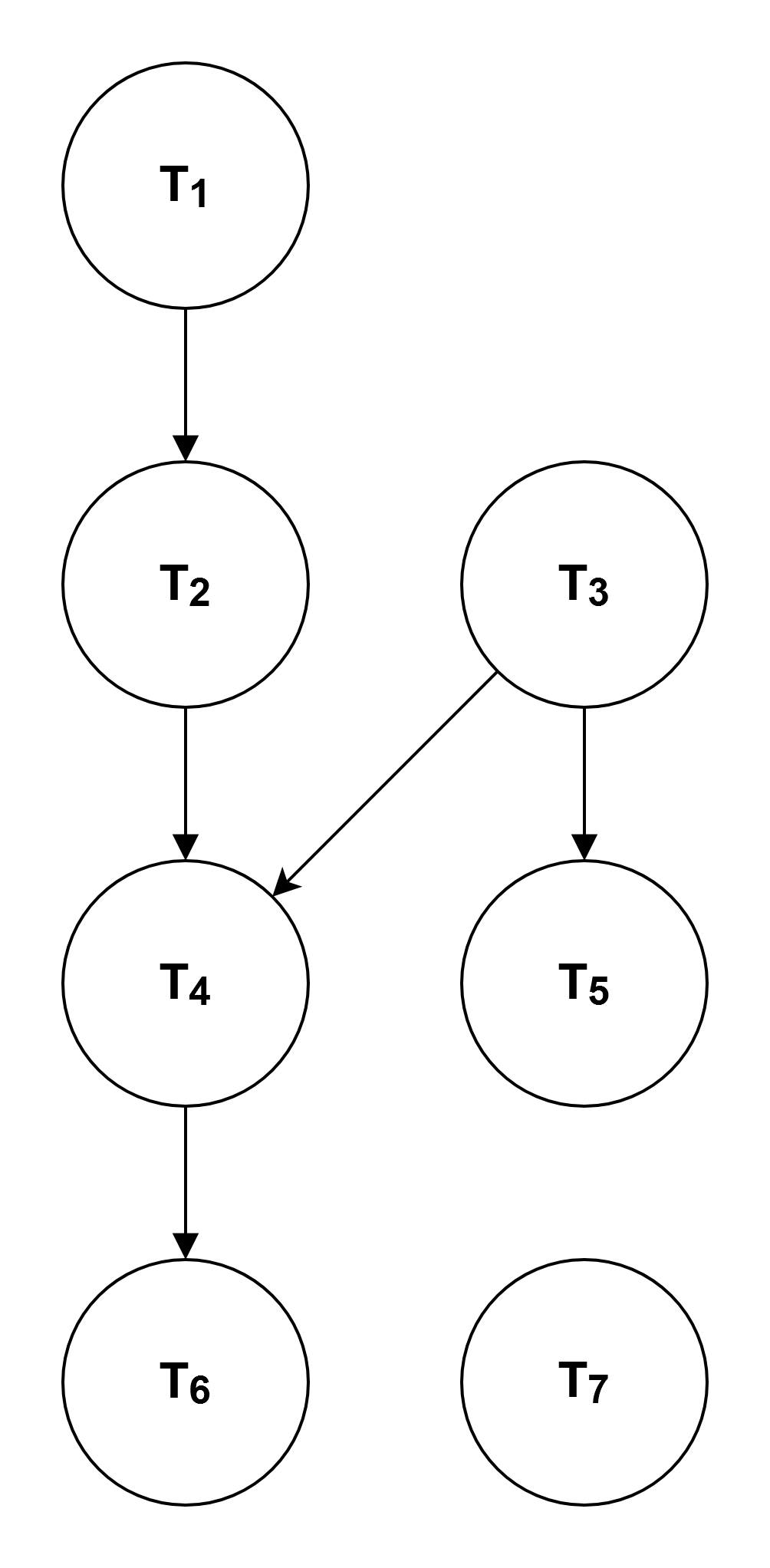}\label{fig:graphbefore}}
  \hspace{0.2cm}
  \subfloat[]{\includegraphics[height=0.2\textheight]{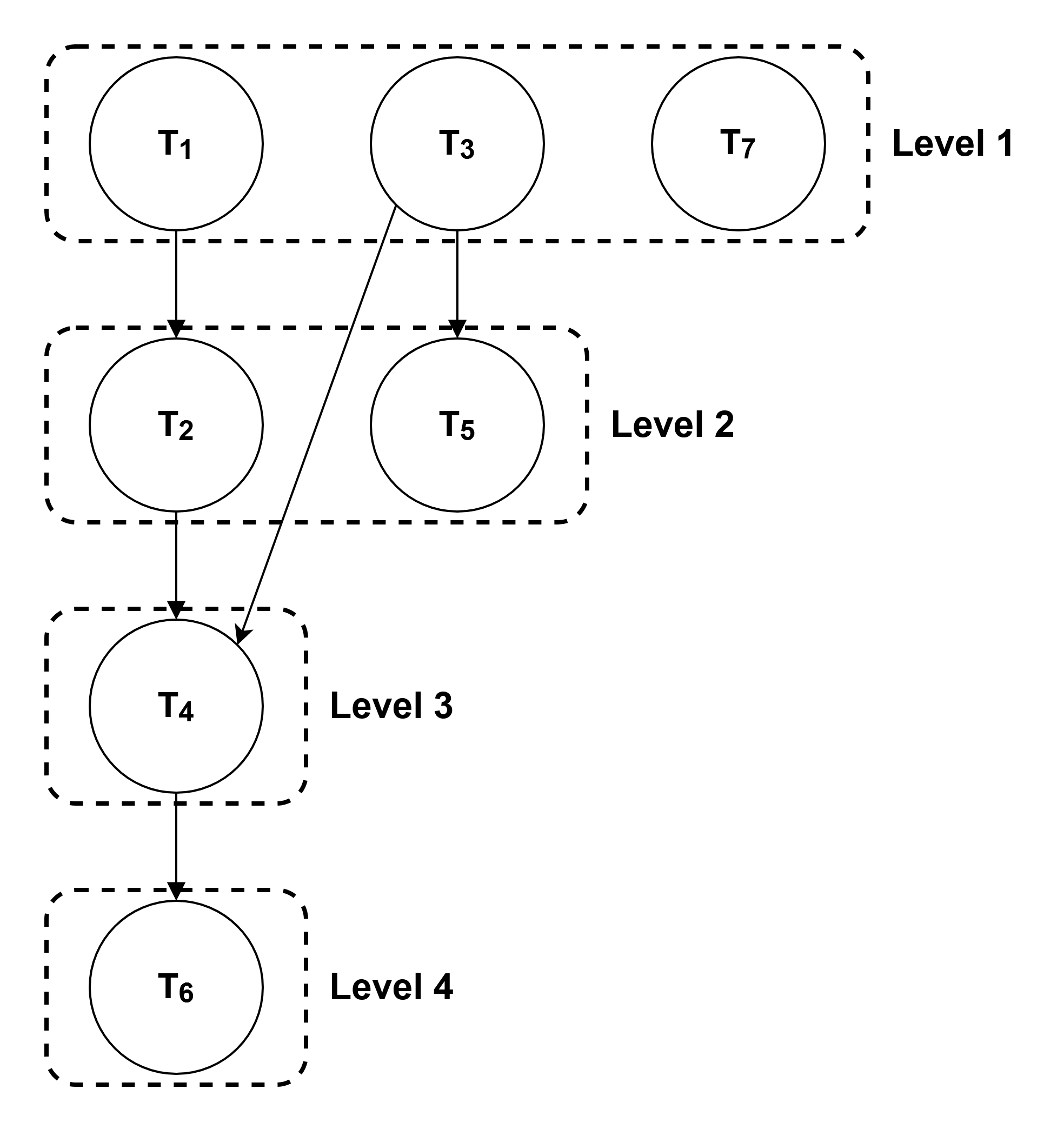}\label{fig:graphafter}}
  \caption{Fig.~\ref{fig:graphbefore} shows the directed acyclic graph of a Job composed by seven tasks, while Fig.~\ref{fig:graphafter} shows the division into four levels.}
\end{figure}

The role of the Task Assignment layer is to build, for each job, the nominal task schedules for the human and the robot, taking into account job quality and precedence constraints. This relation of dependency between tasks can be represented with a directed acyclic graph $G=(T,E)$, as shown in Fig~\ref{fig:graphbefore}. Each vertex represents a task $T_i$ while each directed edge $E_{ij}$ means that the task $T_i$ must be executed before the task $T_j$. Some tasks could be independent of each other, since there is not a path that goes from one task to the other (e.g. $T_1$ and $T_3$). The graph can then be rearranged so that all the parallel tasks are grouped together into several sets called levels $L_l$, as shown in Fig.~\ref{fig:graphafter}. The choice of how the tasks are assigned to each level has a large impact on the schedules.

In this paper, both the problem of allocating the tasks to each agent and the way in which the tasks are distributed over the levels are addressed by solving the following multi-objective Mixed Integer Linear Program:


\begin{equation}
    \label{eq:MILP}
    \begin{array}[l]{ll}
        min_{x,c_l}\sum_{l=1}^L\!\sum_{i=1}^N\!(w_{Ri} x_{Ril}\!+\!w_{Hi} x_{Hil})\!+\!\frac{1}{t_{A,max}}\!\sum_{l=1}^L\!c_l \\\\
        \text{subject to}\\\\
        \sum_{l=1}^L(x_{Ril}+x_{Hil})=1  \quad \forall i \in \{1,\dots , N\}\\\\
        \sum_{i=1}^Nt_{ai} x_{ail}\leq c_l \quad \forall l \in \{1,\dots , L\},\forall a \in A \\\\
        \sum_{a\in A} \sum_{l=1}^L l\cdot x_{ail}< \sum_{a\in A} \sum_{l=1}^L l\cdot x_{ajl} \quad \forall i\rightarrow j \\\\
        K_m \le K_{m,max} \quad \forall m \in \{1,\dots,M\}\\\\
        K_{m,av} \le K_{m,av,max} \quad \forall m \in \{1,\dots,M\}\\\\
    \end{array}
\end{equation}

The terms $w_{Ri}, w_{Hi} > 0 $ represent the weights for executing task $T_i$ on behalf of the robot and of the human, respectively. The Boolean variables $x_{Ril}, x_{Hil} \in \{0,1\}$ are detecting whether $T_i$ is assigned or not to the robot or to the human, respectively, and at what level it must be executed; $x=(x_{R11},\dots, x_{RNL}, x_{H11},\dots, x_{HNL})$ is the vector containing all the decision variables. $t_{ai}>0$ represents the nominal execution time of $T_i$ on behalf of agent $a\in A$ and $t_{A,max}$ is the maximum task duration. $c_l>0$ denotes the cycle time of the $l^{th}$ level and $K_m$ and $K_{m,av}$ are quantitative parameters used to evaluate the $m^{th}$ job quality metric, as detailed in Sec.~\ref{sec:job_quality}. 

$w_{Ri}$ and $w_{Hi}$ are exploited for encoding the cost required by each agent to perform the task (e.g electrical cost, tool wear, or risk assessment). Very high weights are exploited for communicating to the task assignment algorithm that an agent is unsuitable for the execution of a task. Moreover, the human weights are exploited also to embed and evaluate in a quantitative way the job quality. The calculation method of these costs is a design parameter (see e.g. \cite{Lamon2019, liu2020}). In a general way we can define $w_{Ri} = h(costs)$ and \hbox{$w_{Hi} = g(costs, job\,\,quality)$}.

The first constraint guarantees that each task is assigned either to the robot or to the human. The second constraint maximizes the parallelism between the human and the robot. In fact, since all the terms in the quantity to minimize are positive, the optimization problem would tend to choose \hbox{$c~=\sum_{i=1}^Lc_l$} as small as possible and the lower bound for this sum is given by the third constraint and corresponds to the maximum parallelization of the activities of the human and of the robot. The third constraint ensures the respect of the precedence relationship, since all the tasks that should be executed before another are assigned to an upper level. The last constraints impose that the job quality metrics for the human operator will not violate the upper bounds.

The outcome of the optimization problem \eqref{eq:MILP}  are $S_H$ and $S_R$, the set of tasks that have to be executed by the human and by the robot, respectively, and the level at which the task must be executed. This generates the nominal schedules, i.e. two ordered tuples $S_H$ and $S_R$ containing the tasks that have to be sequentially executed by each agent in each level.
\section{Job Quality Metrics for HRC} 
\label{sec:job_quality}
Job quality is considered in the Task Assignment via two mechanisms. The first mechanism ensures that various metrics related to job-quality do not exceed threshold values via optimization constraints. The second mechanism considers the overall attractiveness of the task-set that is assigned to the human. The attractiveness of the task-set is made part of the optimization objective.

Various metrics can be defined that describe a certain task load, e.g. the weight that is lifted, or the noise that is experienced during task execution. To monitor these metrics each task is assigned a set of weights ($k_{i1}, \ldots, k_{iM}$). The metrics ($K_1,\ldots, K_M$) can be calculated using two different generic representation of a job quality metric:


\begin{itemize}
    \item \textit{summed weight:}
    \begin{equation}
        \label{eq:jqmetrics}
        K_m = K_{m,0} + \sum_{l=1}^L\sum_{i=1}^{N}(x_{Hil}k_{im})
    \end{equation}
    \item \textit{average weight:}
    \begin{equation}
        \label{eq:jqmetricsav}
        K_{m,av} = \frac{t_e K_{m,0} + \sum_{l=1}^L\sum_{i=1}^{N}(x_{Hil}t_{Hi}k_{im})}{t_m+c}
    \end{equation}
\end{itemize}
Where $c=\sum_{l=1}^Lc_l$ is the cycle time. Typically, job-quality metrics are calculated over a time span longer than the execution of one task, $t_m$ is the elapsed time of the time frame that is relevant for the metric, e.g. elapsed time since the human started the work shift.  $K_{m,0}$ is the cumulative costs from previous jobs within the time frame $t_e$. The set of cumulative costs of each metric represents the job quality metrics that come out from the Human Monitoring block as shown in Fig.~\ref{fig:framework}. It is worth noting that this cumulative cost guarantees that all the desired job quality metrics are estimated and constrained over the relevant time frames~($t_m$) and not just only over the single schedule.






Ensuring that pre-set thresholds on job-quality aspects are not exceeded does not automatically assign the tasks that are preferred by the human to the human. This aspect is governed by a general  attractiveness factor that is assigned to each task. The attractiveness of the tasks is included in the optimization problem \eqref{eq:MILP}, i.e. in the evaluation of $w_{HI}$, so the scheduler tries to assign those tasks to the human, that the human likes to perform the most.

\section{Dynamic Scheduler} 
\label{sec:dynamic_scheduling}
Starting from the output of the Task Assignment, the goal of the Dynamic Scheduler is to adapt online the two nominal schedules $S_H$ and $S_R$ taking into account the uncertainty of the human behavior. When two humans collaborate, their natural synergy allows them to reach high team performance. If one human gets slower, the other can compensate by speeding up. Furthermore, more complex unexpected difficulties are handled by communication. Experienced team members can exploit their knowledge to reorganize the work or decide to take over a difficult task. Less experienced team members can ask the expert member for some help when problems occur. The dynamic scheduler aims at reproducing this kind of behavior in human-robot collaboration in order to create an effective and natural cooperation.
	
This is achieved exploiting two different strategies. Firstly, the human operator is monitored in real time when performing the task in order to estimate the real execution time and, if necessary, to reschedule the future activities of the robot, reducing waiting time. Secondly, the communication between the human and the robot is enabled, allowing the two agents to take decisions about their activities through the scheduler. In particular, the robot delegates a task it cannot momentarily execute to the human. The human, instead, can decide to execute the task that the robot is performing (because, e.g., from its experience, it knows that the robot is not executing the task properly or to speed up the workflow). Moreover, the human can decide to re-assign some of its tasks to the robot. The dynamic scheduler is implemented according to the pseudo-code reported in Alg.~\ref{alg:dynamicscheduler}.
	
\begin{algorithm}
    \caption{DynamicScheduler()}
    \label{alg:dynamicscheduler}
    \begin{algorithmic}[1]
    	\State \textbf{Require:} $S_H$,$S_R$\label{algl:dsrequire}
    	\State $End_R ,End_H \leftarrow false$ \label{algl:dsinitializeends}
    	\State $l \leftarrow 1$ \label{algl:dsinitializelevel}
    	\If{$S_R(l) \ne \emptyset$ \label{algl:dsinitializetasksstart}} $T_R \leftarrow S_R(l,1)$\EndIf
    	\If{$S_H(l) \ne \emptyset$} $T_H \leftarrow S_H(l,1)$
    	\EndIf \label{algl:dsinitializetasksend}
		\While{$l \leq L$} \label{algl:dswhilelevels}
    	\While{$(T_R\ne\emptyset$ \textbf{and} $T_H\ne\emptyset)$  \label{algl:dswhiletasks}}  	\If{$T_R=\emptyset$ \label{algl:dsrobotidle}}
    	\State {$S_R \leftarrow\mathbf{reschedule}(T_H,S_R)$\label{algl:dsreschedule}}
    	\Else
    	\State {$End_R\leftarrow\mathbf{monitorR}(T_R)$ \label{algl:endr}}
    	\EndIf
    	\State {$End_H\leftarrow \mathbf{checkEndH()}$}
    	\State $M_H\leftarrow \mathbf{read}_H()$, $M_R\leftarrow \mathbf{read}_R()$\label{algl:readmess}
    	\State {$(End_H,End_R,S_H,S_R)=$}
\item[]\hspace{1.5cm}{$\mathbf{communication}(M_H, M_R, S_H,S_R)$} \label{algl:dscommunication}
    	\If{$End_H$}
    	$T_H\leftarrow\mathbf{next} (T_H,S_H(l))$\label{algl:dsnextH}
    	\EndIf
    	\If{$End_R$}
    	$T_R\leftarrow\mathbf{next} (T_R,S_R(l))$\label{algl:dsnextR}
    	\EndIf
    	\EndWhile
    	\State{$l\leftarrow l+1$ \label{alg:dsnextlevel}}
    	\EndWhile
    	\State{$\mathbf{updateJQ()}$\label{alg:dsupdate}}
    \end{algorithmic}
\end{algorithm}

The dynamic scheduler needs as input the nominal task schedules $S_H$ and $S_R$ (Line~\ref{algl:dsrequire}). It firstly sets to false two variables $End_R$ and $End_H$, which identify if the respective agent has concluded its task, and it initializes the job at the first level (Lines~\ref{algl:dsinitializeends}-\ref{algl:dsinitializelevel}). Subsequently, if applicable, the algorithm assigns the first tasks of $S_R(l)$ and $S_H(l)$ to the human and to the robot (Lines~\ref{algl:dsinitializetasksstart}-\ref{algl:dsinitializetasksend}). At this point, it starts two loops to check the end of the job (Line~\ref{algl:dswhilelevels}) and the actual level (Line~\ref{algl:dswhiletasks}), respectively. Inside the second loop, the scheduler first checks if the robot has performed all its tasks in the actual level. If this is true, the robot is in idle, waiting for the human to finish its task, and some tasks may be rescheduled (Line~\ref{algl:dsreschedule}), maximizing the parallelism between the two agents. In the other cases the algorithm exploits the function $\mathbf{monitoR}(T_R)$ to check if the robot has finished the assigned task (Line~\ref{algl:endr}). The $\mathbf{monitorR}()$ function can be implemented using standard procedures, available for robotic applications (see e.g. \cite{pettersson2005}). If the robot cannot succeed to execute $T_R$ (e.g. a timeout error), a $delegate$ message $M_R$ is communicated. Subsequently, the algorithm checks if the human has completed its task, e.g. exploiting an HMI, and all the messages generated by the human and the robot are considered for task swapping (Lines~\ref{algl:dscommunication}). Afterwards, the algorithm checks if the two agents have concluded their tasks and, if it is the case, assigns them the next task in the level (Lines~\ref{algl:dsnextH},~\ref{algl:dsnextR}). If no tasks are scheduled in the actual level, the function $\mathbf{next}(T,S(l))$ returns $\emptyset$. When both $T_R$ and $T_H$ are empty, then the level is concluded and the job moves on to the next one (Line~\ref{alg:dsnextlevel}). Finally, when all the tasks have been performed, the job quality parameters are updated through the function $\mathbf{updateJQ}()$ (Line~\ref{alg:dsupdate}) and used as input for the Task Assignment of the next job.
		
The rescheduling algorithm is represented in Alg.~\ref{alg:reschedule}.
\begin{algorithm}
	\caption{Reschedule()}
	\label{alg:reschedule}
	\begin{algorithmic}[1]
		\State \textbf{Require:} $T_H$, $S_R$ \label{algl:rsrequire}
		\State $t_{res}\leftarrow \mathbf{monitorH}(T_H)$\label{algl:rsmonitor}
		\If{$t_{res}>t_{Ri}$} \label{algl:rscheckresidualtime}
		\State $(pS_R,fS_R) \leftarrow \mathbf{split}(T_R,S_R)$\label{algl:rssplit}
		\State $fS_R^r \leftarrow\mathbf{fill}(fS_R,t_{res}-t_{Ri})$\label{algl:rsknap}
		\State $S_R \leftarrow\mathbf{concat}(pS_R. S_R^r,fS_R/fS_R^r)$\label{algl:rsconcat}
		\EndIf
		\State $\mathbf{return}$ $S_R$\label{algl:rsreturn}
	\end{algorithmic}
\end{algorithm}	
The algorithm requires as input the task $T_H$ is currently assigned to the human and the current robot schedule $S_R$ (Line~\ref{algl:rsrequire}). It exploits the human monitoring strategy to estimate the remaining time $t_{res}$ for the accomplishment of $T_H$ (Line~\ref{algl:rsmonitor}). If $t_{res}$ is greater than the time necessary for the execution of some tasks in $S_R$, then these tasks may be executed in parallel with $T_H$ and, therefore, the rescheduling procedure starts (Line~\ref{algl:rscheckresidualtime}). First, the schedule $S_R$ is split into two sub-lists: $pS_R$ contains all the robot assigned tasks in the actual level while $fS_R$ contains all the tasks of the next levels, which still need to be performed (Line~\ref{algl:rssplit}). $fS_R$ is then used to create another sub-list $S_R^r$, which contains all the tasks that can be executed in the extra time available $t_{res}-t_{Ri}$ and whose precedences have already been executed (Line~\ref{algl:rssplit}). Subsequently, $S_R(l)$ is updated by  concatenating $pS_R$, with the list of the rescheduled tasks $fS_R^r$ (Line~\ref{algl:rsconcat}). Finally, the new schedule $S_R$ is returned.

Several implementation of procedure $\mathbf{monitorH}()$ are available in the literature as, e.g., \cite{vo2014,maeda2017,maderna2019}.

\begin{figure}[t]
\vspace*{5pt}
	\centerline{\includegraphics[width=\columnwidth]{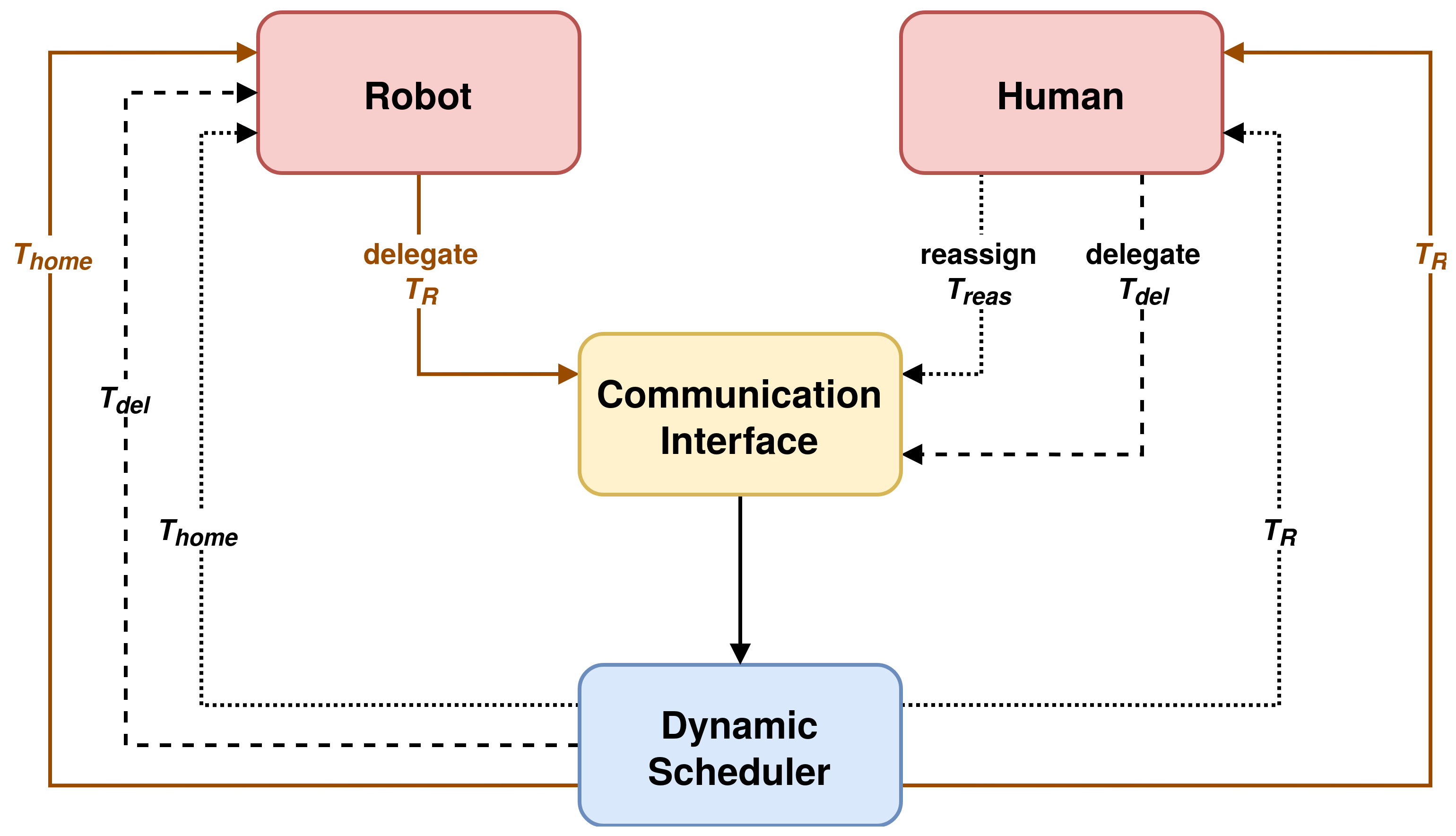}}
	\caption{Communication Layout. The dotted lines indicate the ``reassign" message coming from the human and the consequent task scheduling. The dashed lines indicate the ``delegate" message coming from the human with the following scheduling to the robot. Finally, the brown lines indicate the ``delegate" message communicated by the robot.}
	\label{fig:messages}
\end{figure}

During the execution of the job the human and the robot can generate messages in order to communicate to the Dynamic Scheduler the intention or need to swap their tasks. A detailed representation of this communication layout is shown in Fig.~\ref{fig:messages}.
In particular, the message $M_R$ sent by the robot can be either empty or containing the value ``delegate $T_{R}$''  and it is generated by the \textbf{monitorR} function when the robot cannot succeed in executing the assigned task. The message $M_H$ can be either empty or it can assume two values: ``reassign $T_{reas}$'' or ``delegate $T_{del}$''. The first message is generated when the human decides to execute the task that the robot is executing (because, e.g., the robot is not doing the assigned work properly or in the best way). The second message is generated when the human decides to delegate some tasks in $S_H$ to the robot. This message has an argument, that specifies the task to be delegated. The human can enter the messages through a proper, job dependent, input interface. The messages generated by the human and by the robot are handled by Alg.~\ref{alg:communication}.
	
\begin{algorithm}
	\caption{Communication()}
	\label{alg:communication}
	\begin{algorithmic}[1]
		\State \textbf{Require:} $M_H$, $M_R$, $T_R$, $S_H$,$S_R$ \label{algl:corequire}
		\If{$M_H=reassign(T_{reas})$ } 
		\State $End_H\leftarrow true$\label{algl:coendh}
		\If{$T_{reas} = T_R$} \label{algl:cochecktreas}
    		\State $End_R\leftarrow true$\label{algl:coendr}
			\State$S_R\leftarrow\mathbf{push}(T_{home},S_R)$\label{algl:copushhome1}
		\EndIf
		\State
		$S_R\leftarrow\mathbf{delete}(T_{reas},S_R)$\label{algl:codelete3}
		\State  $S_H\leftarrow\mathbf{push}(T_{reas},S_H)$\label{algl:copushRH}
		\ElsIf{$M_H=delegate(T_{del})$ \textbf{and} $\mathbf{exRobot}(T_{del})$}\label{algl:cocheckexrob}
		\State $End_H\leftarrow true$\label{algl:coendh2}
		\State $S_H\leftarrow\mathbf{delete}(T_{del},S_H)$\label{algl:codelete1}
		\State   $S_R\leftarrow\mathbf{push}(T_{del},S_R)$\label{algl:copushdelR}
		\EndIf
		\If{$M_R=delegate(T_{R})$ \textbf{and} $\mathbf{exHuman}(T_R)$}\label{algl:cocheckexhum}
		\State $End_R\leftarrow true$\label{algl:coendr2}
		\State $S_R\leftarrow\mathbf{delete}(T_R,S_R)$\label{algl:codelete2}
		\State $S_R\leftarrow\mathbf{push}(T_{home},S_R)$\label{algl:copushhomeR}
		\State   $S_H\leftarrow\mathbf{push}(T_R,S_H)$\label{algl:copushRH2}
		\EndIf
		\State $\mathbf{return}$ $End_H$, $End_R$, $S_H$, $S_R$\label{algl:coreturn}
	\end{algorithmic}
\end{algorithm}	

The algorithm requires the messages generated by human and robot $M_R$ and $M_H$, the current schedules $S_R$ and $S_H$ and the task $T_R$ currently assigned to the robot (Line~\ref{algl:corequire}). The message $M_H$ is the first to be handled in order to give priority to the decisions taken by the operator. If the human decides to execute a task that was initially assigned to robot, it is necessary to check if the robot already started this task. If this is true, the robot task execution is aborted, moving the robot in an home safe position. $T_{reas}$ is then deleted from $S_R$ and pushed in the first position of the human schedule \hbox{(Lines~\ref{algl:coendh}-\ref{algl:copushRH})}. The human can also decide to delegate a task $T_{del}\in S_H$. If $T_{del}$ is executable by the robot, it is deleted from $S_H$ and transferred into the robot schedule  \hbox{(Lines~\ref{algl:cocheckexrob}-\ref{algl:copushdelR}).} 
The robot, instead, could detect that it cannot fulfill the assigned task and, if the task is executable by the human operator, $T_R$ is deleted from the robot schedule and inserted in the schedule of the human, while a homing mission $T_{home}$ is added as the next task for the robot \hbox{(Lines \ref{algl:cocheckexhum}-\ref{algl:copushRH2})}.
Finally the procedure returns the updated end of task variables and schedules (Line~\ref{algl:coreturn}).
The procedures \textbf{exRobot} and \textbf{exHuman} exploit prior information about the job and the tasks (e.g. the weights $w_{Ri}$ and $w_{Hi}$  in \eqref{eq:MILP}) to detect if a task can be executed by the robot or by the human.

Swapping the tasks between the human and the robot directly affects the job quality. Thus, it may happen that the final job quality indices do not respect the constraint imposed in the MILP problem \eqref{eq:MILP}. However, since the optimization problem requires as input the real job quality parameters, namely $K_{m,0}$, the possible work overload for the operator will be mitigated with subsequent task schedules. Another possible way to ensure the optimum of the job quality is the implementation of a function that checks if the task swapping will violate the constraints and, if required, prevents it. However the latter strategy is very conservative, as it does not take into account the possibility of correcting the job quality parameters in subsequent optimization problems. Furthermore, at the communication level, the job quality is treated as a soft constraint with respect to the need to swap a task (e.g. the robot failing a task is a more critical situation). For these reasons, in this paper it was decided not to investigate the latter solution. 

The Dynamic Scheduler is computationally cheap. The heaviest part is represented by the $\mathbf{fill()}$ function inside the $\mathbf{Reschedule()}$ algorithm (see Alg.~\ref{alg:reschedule}, Line \ref{algl:rssplit}). This function, in its worst implementation, has a linear complexity equal to $O(n+l)$, i.e. when analyzing every single element of the tuple with two \textit{for} loops. Since the number of tasks and levels in an industrial application is not that large, the algorithm is reactive. 

\section{Experiments} 
\label{sec:experiments}
\begin{figure}[!tbp]
	\vspace*{5pt}
		\centering
		\begin{minipage}[b]{0.49\columnwidth}
			\includegraphics[width=\columnwidth]{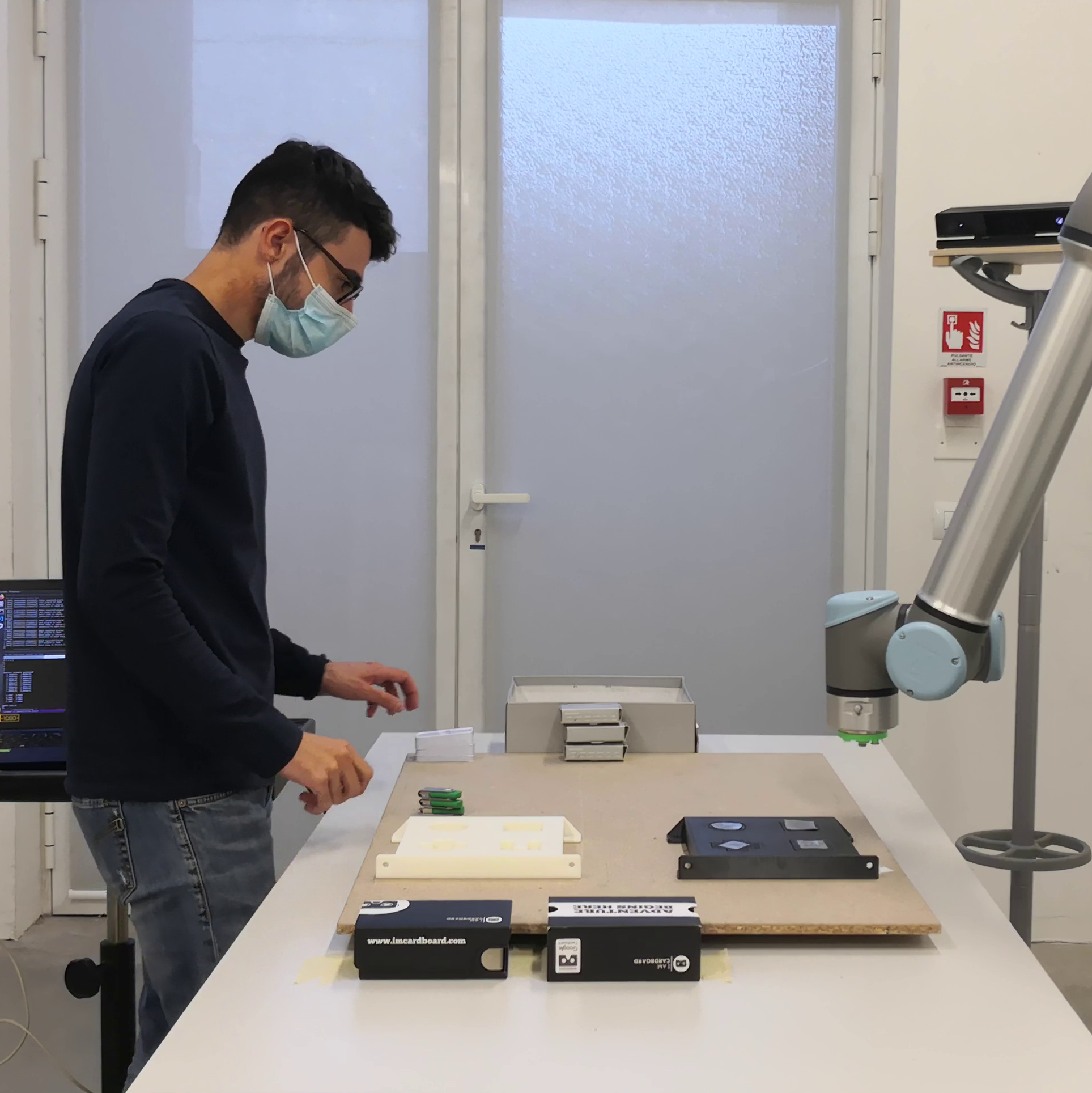}
		\end{minipage}
		\hfill
		\begin{minipage}[b]{0.49\columnwidth}
			\includegraphics[width=\columnwidth]{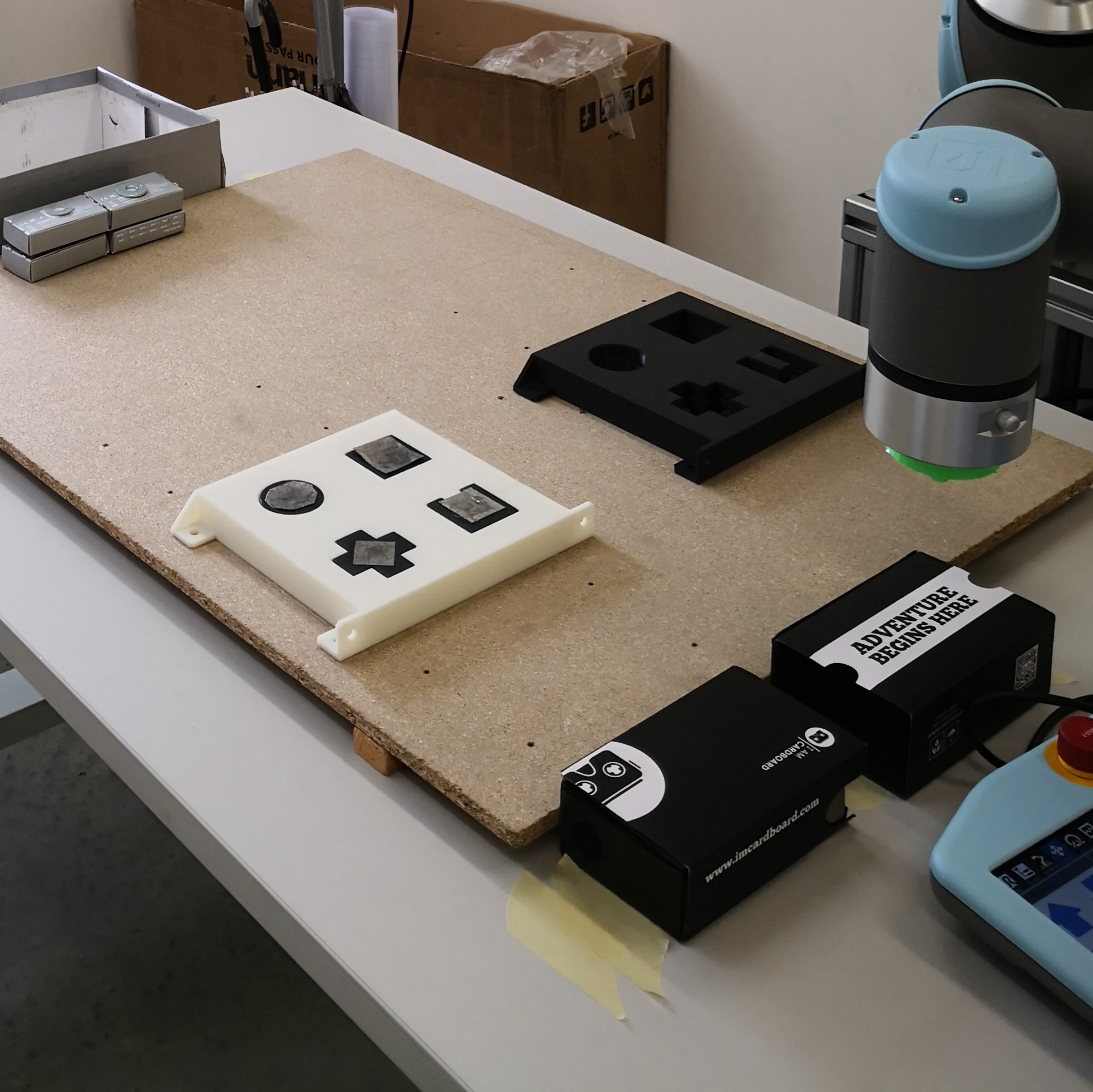}
		\end{minipage}
		\caption{Setup of the experiment. The two images show all the equipment used during the experiments.}
		\label{fig:photo_setup}
	\end{figure}

The proposed two-layered framework has been experimentally validated in a custom collaborative assembly work shift, that has been set up only for evaluation purpose. During the experiments, the human operator cooperated with a UR10e, a 6-DoF collaborative robot. 
To monitor the human task execution we used a Kinect V2 RGB-D Camera with the official APIs for the skeleton tracking and to evaluate the remaining task time we implemented the OE-DTW algorithm, which is available in literature (see \cite{tormene}), to both operator wrists. This algorithm returns the percentage completion of the task $\%_{compl}$, comparing the actual time series with the reference ones. This percentage was then exploited to estimate the remaining time of the task as $t_{res_i} = (1-\%_{compl})t_{Hi}$. For the communication interface we used the computer keyboard to send signal through a simple HMI. The complete setup for the experiment is shown in Fig.~\ref{fig:photo_setup}.

All the software components were developed using ROS Melodic Morenia.
The optimization problem was implemented using Python-MIP~\cite{mip},
a collection of Python tools for the modeling and solution of Mixed-Integer Linear programs, and solved with Gurobi solver~\cite{gurobi}. The UR10e is position controlled using the ROS interface which accepts a desired final position while the overall trajectories are directly generated by the low level controller.
		
The work shift was composed by two jobs divided into multiple tasks. These are listed in Tab.~\ref{tab:ds}.

\begin{table}[t]
    \vspace{3pt}
	\centering
	\caption{Tasks Description}
	\label{tab:ds}
	\newcolumntype{Y}{>{\centering\arraybackslash}X}
	\newcolumntype{s}{>{\hsize=.5\hsize}Y}
	\begin{tabularx}{\columnwidth}{sYs}
		\toprule
		\textbf{Task Index} & \textbf{Description} & \textbf{Job}\\
		\midrule
		1 & Pick\&Place square shape. & $J_1$, $J_2$\\
		\midrule
		2 & Pick\&Place U shape. & $J_1$, $J_2$\\
		\midrule
		3 & Pick\&Place circular shape. & $J_1$, $J_2$\\
		\midrule
		4 & Pick\&Place cross shape. & $J_1$, $J_2$\\
		\midrule
		5 & Pick\&Place weight. & $J_1$, $J_2$\\
		\midrule
		6 & Pick\&Place weight. & $J_1$, $J_2$\\
		\midrule
		7 & Packaging USB key. & $J_1$\\
		\midrule
		8 & Packaging USB key. & $J_1$\\
		\midrule
		9 & Packaging USB key. & $J_1$\\
		\bottomrule
	\end{tabularx}
\end{table}

In the experiments we considered the average weight the human has to lift during the execution of the job $K_{1,av}$ as a job quality metric, with its respective upper bound $K_{1,av,max} = 1.1$.

The other inputs required by the Task Assignment Layer (see Sec.~\ref{sec:task_planning}) are shown in Tab.~\ref{tab:optimization_data}. 
\begin{table}[t]
	\centering
	\caption{Task Assignment Data}
	\label{tab:optimization_data}
	\newcolumntype{Y}{>{\centering\arraybackslash}X}
	\begin{tabularx}{\linewidth}{YYYYYYY}
		\toprule
		\textbf{Task Index} & $\bm{w_{Ri}}$ & $\bm{t_{Ri}}$ & $\bm{w_{Hi}}$ & $\bm{t_{Hi}}$ & $\bm{k_{i1}}$ & $\bm{i\rightarrow j}$\\
		\midrule
		1 & 0.1 & 12  & 0.4 & 15 & 0 & -\\
		\midrule
		2 & 0.1 & 12 & 0.4 & 15  & 0 & -\\
		\midrule
		3 & 0.1 & 12 & 0.4 & 15  & 0 & 1, 2\\
		\midrule
		4 & 0.1 & 12 & 0.4 & 15  & 0 & 1, 2\\
		\midrule
		5 & 0.5 & 25  & 0.8 & 10 & 9 & -\\
		\midrule
		6 & 0.5 & 25 & 0.8 & 10  & 9 & -\\
		\midrule
		7 & 1000 & - & 0.4 & 25 & 0 & -\\
		\midrule
		8 & 1000 & - & 0.4 & 25 & 0 & -\\
		\midrule
		9 & 1000 & - & 0.4 & 25 & 0 & - \\
		\bottomrule
	\end{tabularx}
\end{table}
The nominal durations were estimated by computing the average value of multiple measurements, while the intrinsic costs were calculated with the following equations:
\begin{align}
    w_{Ri} &=0.7D_{Ri} + 1000(1-capability_i) \\
    w_{Hi} &= u_i
\end{align}
where $D_{Ri}$ represents the distance that the robot has to perform during the execution of the task $T_i$ and $capability_i$ is a Boolean variable that indicates if the robot is capable of perform the task, e.g. the robot is not able to execute tasks $T_7,T_8,T_9$. $u_i$ is the attractiveness factor for the human in executing the task $T_i$ and it was defined using our experience and knowledge in the work. $k_{i1}$ are defined taking into account the weight of the objects, i.e. holding an USB stick or a shape does not affect the analyzed job quality metric. The only precendence constraints are related to the shapes and they are necessary to avoid the robot making a wrong pick with the magnets.

When the work shift starts. the job $J_1$ must be executed and the Task Assignment layer is initialized with the cumulative cost $K_{1,0} = 0$. The optimization problem is solved in $100\,ms$ and the schedule is composed by the following two tuples:
\begin{itemize}
	\item $S_H = \{[7,8,5], [9]\}$
	\item $S_R = \{[3,4,6], [1,2]\}$
\end{itemize}
with an estimated job quality metric equals to:
\begin{equation}
    K_{1,av}=\frac{t_{H5}k_{15}}{c} = 1.1
\end{equation}

Starting from the output of the Task Assignment Layer, the Dynamic Scheduler was then initialized and the two agents began to perform the collaborative job. A video of the experiments can be found in the supplementary material. The first part of the video is dedicated to the execution of the \textit{``nominal schedule"} of the job $J_1$, i.e. the two agents perform exactly the assigned tasks. When the robot concludes all the tasks of the first level, the Dynamic Scheduler exploits the information coming from the OE-DTW to reschedule some tasks (see Alg.~\ref{alg:reschedule}, Line~\ref{algl:rscheckresidualtime}~--~\ref{algl:rsconcat}). Since the monitoring algorithm returns an estimated remaining time $t_{res}>t_{R1}$, the robot can anticipate the task $T_1$ in the first level, instead of waiting for the level to finish as scheduled. This procedure is executed in $4\,ms$. 

Once $J_1$ is concluded the real execution time $t_{H5} = 15\,s$ and the real duty cycle $c = 79\,s$ are exploited to calculate the real job quality metric $K_{1,av} = 1.7$.
The real job quality metric is then used as input for the assignment procedure of the next job $J_2$. The new optimization problem is solved in $90\,ms$ and the resulting schedules are:
\begin{itemize}
	\item $S_H = \{[3], [1, 2]\}$
	\item $S_R = \{[4], [5, 6]\}$
\end{itemize}

It is worth noting that thanks to the job quality constraint no tasks affecting the weight metric are assigned to the human and the new estimated job quality metric is below the upper bound: 
\begin{equation}
    K_{1,av} = \frac{K_{m,0}}{t_e+c} = 0.96 \le 1.1
\end{equation}

The second part of the video shows the execution of $J_2$.

The communication strategy is then exploited. The work shift is reinitialized and the operator starts to execute the nominal schedule of the job $J_1$. After concluding $T_7$ and $T_8$ he sends a message ``delegate $T_5$" to the Dynamic Scheduler and the robot starts to execute this task instead of the human. At the end of the schedule, the human asks to the Dynamic Scheduler to reassign the task $T_2$, and he performs that task instead of the robot.
Since the human did not perform any task that affects the weight metric, the cumulative cost for the assignment of the $J_2$ is $K_{1,0}=0$. For this reason, the output of the Task Assignment for $J_2$ is:
\begin{itemize}
	\item $S_H = \{[\,], [6]\}$
	\item $S_R = \{[3,4], [1,2,5]\}$
\end{itemize}
with an estimated weight metric $K_{1} = 1$. This solution is obtained in $100\,ms$. It is worth noting that $c=90\,s$ this is due to the fact that the Task Assignment schedules a pause before starting the execution of the second level. This pause is necessary to obtain an admissible value for the weight metric.

In order to demonstrate the effectiveness of the architecture, $J_1$ is performed without rescheduling the tasks. As happened before, while the human performs its tasks the robot concludes the first level as planned. Since the rescheduling is not active, the robot stops, waiting for the human to conclude the first level of the schedule. After the human executes $T_5$, the schedule passes to the second level and two agents resume the expected behavior.
As expected, the trial without the rescheduling takes more time ($c = 85\,s$). Moreover, a great improvement can be seen in the robot idle times: $T_{R,idle}=12\,\,s$ with our framework and $T_{R,idle}=20\,\,s$ without using the rescheduling procedure.
\section{Conclusions} 
\label{sec:conclusions}
In this paper we have proposed a two layer architecture for dynamic task assignment and scheduling for collaborative application. Job quality has been explicitly considered. Rescheduling is performed considering real-time monitoring of the operator's activities and the human robot communication. The framework has been tested in a a custom HRC assembly scenario and the results seem good, since the expectations have been satisfied.

Future work aims at validating the proposed architecture in an industrial scenario, where a real shift is considered and a user study is implemented. This will lead to a further validation of the architecture potential. Moreover, the HMI could be substituted with a different communication system that makes the message exchange simpler and more intuitive. Finally, how the job quality is affected by the rescheduling and the communication system could be analyzed, improving even more the well-being of the human operator in HRC.


%
%
\bibliographystyle{IEEEtran}
\normalem
\bibliography{bibliography}
\end{document}